\documentclass{article} 
\usepackage{iclr2015,times}
\usepackage{hyperref}
\usepackage{url}
\usepackage{amsmath}
\usepackage{amssymb}
\usepackage{graphicx}
\usepackage[font=small]{caption}

\title{Self-informed neural network \\ structure learning}

\author{
David Warde-Farley \thanks{Work done while \textsc{DWF} was an intern at Google.} \\
D\'epartement d'Informatique et de Recherche Op\'erationelle \\
Universit\'e de Montr\'eal \\
Montreal, Quebec, Canada \\
\texttt{wardefar@iro.umontreal.ca} \\
\And
Andrew Rabinovich \& Dragomir Anguelov \\
Google, Inc. \\
Mountain View, CA 94043, USA \\
\texttt{\{amrabino,dragomir\}@google.com} \\
}

\iclrfinalcopy 

\begin{document}

\maketitle

\begin{abstract}
We study the problem of large scale, multi-label visual recognition with a
large number of possible classes. We propose a method for augmenting a trained
neural network classifier with auxiliary capacity in a manner designed to
significantly improve upon an already well-performing model, while minimally
impacting its computational footprint. Using the predictions of the network
itself as a descriptor for assessing visual similarity, we define a
partitioning of the label space into groups of visually similar entities. We
then augment the network with auxilliary hidden layer pathways with
connectivity only to these groups of label units. We report a significant
improvement in mean average precision on a large-scale object
recognition task with the augmented model, while increasing the number of
multiply-adds by less than 3\%.  \end{abstract}

\section{Introduction}
In the context of large scale visual recognition, it is not uncommon for
state-of-the-art convolutional networks to be trained for days or weeks before
convergence~\citep{krizhevsky2012imagenet,sermanet-iclr-14,Szegedy14arXiv}. Performing exhaustive architecture search is quite challenging and
computationally expensive. Furthermore, once a satisfactory architecture has
been discovered, it can be extremely difficult to improve upon; small changes
to the architecture more often decrease performance than improve it. In
architectures containing fully-connected layers, naively increasing the
dimensionality of such layers increases the number of parameters between them
quadratically, increasing both the computational workload and the tendency
towards overfitting.

In settings where the domain of interest comprises thousands of classes,
improving performance on specific subdomains can prove challenging, as the
jointly learned features that succeed on the overall task on average may not be
sufficient for correctly identifying the ``long tail'' of classes, or
for making fine-grained distinctions between very similar entities. Side
information in the form of metadata -- for example, from Freebase~\citep{Freebase} -- often only
roughly corresponds to the kind of similarity that would make correct
classification challenging. In the context of object classification, visually
similar entities may belong to vastly different high-level categories (e.g. a
sporting activity and the equipment used to perform it), whereas two entities
in the same high-level semantic category may bear little resemblance to one
another visually.

A traditional approach to building increasingly accurate classifiers is to
average the predictions of a large ensemble. In the case of neural networks, a
common approach is to add more layers or making existing layers significantly
larger, possibly with additional regularization. These strategies
present a significant problem in runtime-sensitive
production environments, where a classifier must be rapidly evaluated in a
matter of milliseconds to comply with service-level agreements.  It is
therefore often desirable to increase a classifier's capacity in a way that
significantly improves performance while minimally impacting the computational
resources required to evaluate the classifier; however, it is not immediately
obvious how to satisfy these two competing objectives.

We present a method for judiciously adding capacity to a trained neural network
using the network's own predictions on held-out data to inform the augmentation
of the network's structure. We demonstrate the efficacy of this method by using
it to significantly improve upon the performance of a state-of-the-art
industrial object recognition pipeline based on \cite{Szegedy14arXiv} with
less than 3\% extra computational overhead.

\section{Methods}

Given a trained network, we evaluate the network on a held out dataset
in order to compute a confusion matrix. We then apply spectral
clustering \citep{chung1997spectral} to generate a partitioning
of the possible labels.

We augment the trained network's structure by adding additional stacks of fully
connected layers, connected in parallel with the pre-existing stack of
fully-connected layers. The output of each ``auxiliary head'' is connected by a
weight matrix only to a subset of the output units, corresponding to the
label clusters discovered by spectral clustering.

\begin{figure}
\centering
\includegraphics[width=5in]{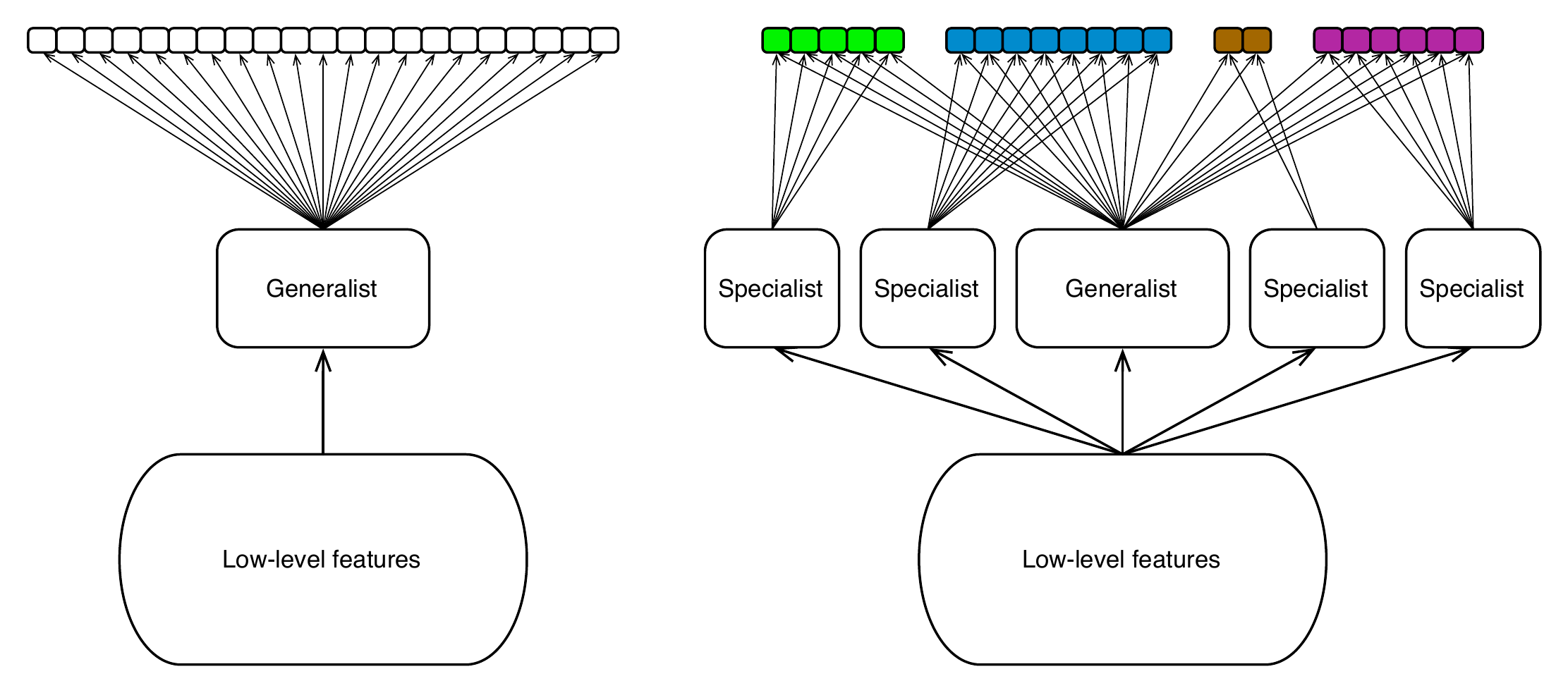}
\caption{A schematic of the augmentation process. Left: the original network.
Right: the network after augmentation.}
\end{figure}

We train the augmented network by initializing the pre-existing portions of the
network (minus the classifier layer's weights and biases) to the parameters of
the original network, and by randomly initializing the remaining portions. We
train holding the pre-existing weights and biases fixed, learning only the
hidden layer weights for the new portions and retraining the classifier layer's
weights. This allows for training to focus on making good use of the auxiliary
capacity rather than adapting the pre-initialized weights to compensate for the
presence of the new hidden units. Note that it is also possible to fine-tune
the whole network after training the augmented section, though we did not
perform such fine-tuning in the experiments described below.

\section{Related work}

Our method can be seen as similar in spirit to the mixture of experts approach
of \cite{jacobs1991adaptive}. Rather than jointly learning a gating function
as well as experts to be gated, we employ as a starting point a strong
generalist network, whose outputs then inform decisions about which specialist
networks to deploy for different subsets of classes. Our specialists also do
not train with the original data as input but rather a higher-level feature
representation output by the original network's convolutional layers.

Recent work on distillation~\citep{Hinton2014}, building on
earlier work termed model compression~\citep{Bucilua2006},
emphasizes the idea that a great deal of valuable information can be gleaned
from the non-maximal predictions of neural network classifiers. Distillation
makes use of the averaged overall predictions of several expensive-to-evaluate
neural networks as ``soft targets'' in order to train a single
network to both predict the correct label and mimic the overall predictions of
the ensemble as closely as possible. As in \cite{Hinton2014}, we use the
predictions of the model itself, however we use this knowledge in the pursuit
of carefully adding capacity to a single, already trained network, rather than
mimicking the performance of many networks with one. Our approach is arguably
complementary, and could conceivably be applied after distilling an ensemble
into a single mimic network in order to further improve fine-grained
performance.

\section{Experiments}
Our base model consists of the same convolutional Inception architecture
employed in GoogLeNet~\citep{Szegedy14arXiv}, plus two fully
connected hidden layers of 4,096 rectified linear (ReLU) units each. Our output
layer consists of logistic units, one per class.

We evaluated the trained network on 9 million images not used during training.
Let \begin{align}
g_j(x) & = \left\{\begin{array}{cl}
1, & \text{if example $x$ has ground truth annotation for class $j$} \\
0, & \text{otherwise}
\end{array}
\right. \\
M_{i,K}(x) & = \left\{\begin{array}{cl}
1, & \text{if model $M$'s top $K$ predicted labels on example $x$ includes class $i$} \\
0, & \text{otherwise}
\end{array}
\right.
\end{align}
We compute the following matrix on the hold-out set $S$:
\begin{align}
\label{eqn:confusion}
A = [a_{ij}];\ \  a_{ij} = \mathbb{E}_{x \in S} \left[ M_{i,K}(x) \cdot g_j(x) \right]
\end{align}
using $K = 100$. We use the seemingly large value of $K = 100$ in order to
recover annotations for a large fraction of possible classes on at least one
example in the hold-out set.  We term the detection of class $i$ in the context
of ground truth class $j$ a \textit{confusion} of $i$ with $j$; the $(i, j)$th
entry of this matrix thus encodes the fraction of the time class $i$ is 
``confused'' with class $j$ on the hold-out set.

We also experimented with the matrix
\begin{align}
\label{eqn:codetection}
A = [a_{ij}];\ \  a_{ij} = \mathbb{E}_{x \in S} \left[ M_{i,K}(x) \cdot M_{j, K}(x) \right]
\end{align}
wherein we eschew the use of ground truth and only look at
\textit{co-detections}, again with $K = 100$.

We symmetrize either matrix as $B = A^\mathsf{T}A$, and apply spectral
clustering using $B$ as our similarity matrix, following the formulation
of \cite{Ng2002}. In all of our experiments, our specialist sub-networks
consisted of two layers of 512 ReLUs each.



We evaluate our method on an expanded version of the JFT dataset described
in~\cite{Hinton2014}, an internal Google dataset with a training set of
approximately 100 million images spanning 17,000 classes.

\section{Results}

\subsection{Label clusters recovered}

In Table \ref{table:clusters}, we observe that spectral clustering on the
matrix $B$ was able to successfully recover clusters consisting of visually
similar entities.
\begin{table}[h]
\centering
\tiny{
\begin{tabular}{c}
\hline
\\
Runway, Handshake, Douglas dc-3, Tarmac, Boeing, Air show, Interceptor, Hospital ship, Coast guard, Republic p-47 thunderbolt,\\
Sikorsky sh-3 sea king, Boeing 737, Mcdonnell douglas dc-10, Air force, Boeing
757, Boeing 717, Hovercraft, Lockheed ac-130, McDonnell Douglas, Travel, \\
Aircraft engine, Flight, Yawl, Lockheed c-5 galaxy, Cockpit, Bomber, Lockheed
p-3 orion, Avro lancaster, Jet aircraft\ldots \\ \\ \hline
\\
Pickled food, Grilled food, North african cuisine, Vinegret, Woku, Lasagne,
Lard, Meringue, Peanut butter and jelly sandwich, Sparkling wine, Salting, \\
Raclette, Mussel, Galliformes, Chemical compound, Succotash, Cucurbita,
Alcoholic beverage, Bento, Osechi, Okonomiyaki, Nabemono, Miso soup, Dango, \\
Onigiri, Tempura, Mochi, Soba, Shiitake, Indian cuisine, Andhra food, Foie gras, Krill, Sour cream, Saumagen, Compote\ldots \\ \\ \hline
\\
Lingonberry, Rooibos, Persimmon, Rutabaga, Banana family, Ensete, Apple, Viola,
Shamrock, Walnut, Beech, Poppy, Kimjongilia, Chicory, Bay leaf, \\
Melon, Grain, Juniper, Spruce, Fir, Birch family, Hawthorn, Guava, Gooseberry,
Tick, Pouchong, Bonsai, Caraway, Fennel, Sea anemone, Maple sugar,\\
Boysenberry, Mustard and cabbage family, Pond, Moss, Daikon, Wild ginger,
Groundcover, Holly, Viburnum lentago, Ivy family, Mustard seed\ldots \\
\\ \hline
\end{tabular}
}
\caption{Examples of partial sets of labels grouped together by performing
spectral clustering on the base network's confusions, based on the 100 top
scoring predictions. The first row appears aviation-related, the second
focusing on mainly food, and the third broadly concerned with plant-related
entities.}
\label{table:clusters}
\end{table}
\subsection{Test set performance improvements}

We evaluate on a balanced test set with the same number of classes per image.
For each of the confusion and co-detection cases, we compare against a network
with identical capacity and topology (i.e. same number of labels per cluster)
with labels randomly permuted, in order to assess the importance of the
particular partitioning discovered while carefully controlling for the number
of parameters being learned.

\begin{center}
\begin{tabular}{l|c|c|c}
Description & mAP @ top 50 & \# Multiply-Adds & Extra Computation \\ \hline
Base network & 36.80\% & 1.52B & 1.000$\times$ \\ \hline
Base + 6 heads, confusions & 39.41\% & 1.56B & 1.026$\times$ \\
Base + 6 heads, randomized & 32.97\% & '' & '' \\ \hline
Base + 13 heads, co-detections & 38.07\% & 1.60B & 1.053$\times$ \\
Base + 13 heads, randomized & 32.13\% & '' & ''
\end{tabular}
\end{center}
While both methods improve upon the base network, the use of ground truth
appears to provide a significant edge. Our best performing network, with 6
specialist heads, increases the number of multiply-adds required for evaluation
from 1.52 billion to 1.56 billion, a modest increase of 2.6\%.

\begin{figure}[h]
\centering
\includegraphics[width=3in]{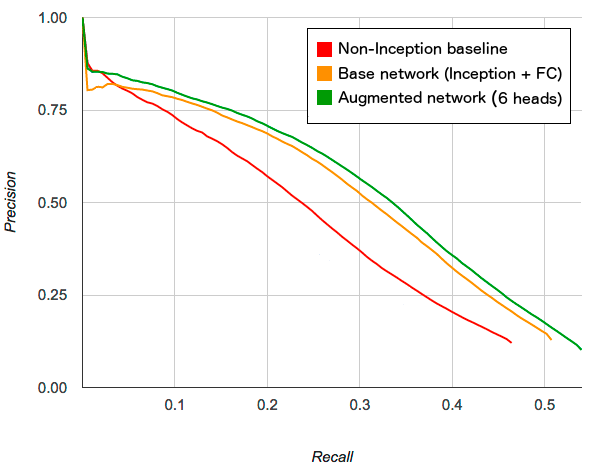}
\caption{A preliminary evaluation of our trained network on the subset of classes in JFT that are mappable to the 1,000-class ImageNet classification task.}
\label{fig:imagenet}
\end{figure}

We also provide, in Figure~\ref{fig:imagenet}, an evaluation of our best
performing JFT network against the ImageNet 1,000-class test set, on the subset
of JFT classes that can be mapped to classes from the ImageNet task
(approximately 660 classes). These results are thus not directly comparable to
results obtained on the ImageNet training set; a more direct comparison is left
to follow-up work.

\section{Conclusions \& Future Work}

We have presented a simple and general method for improving upon trained
neural network classifiers by carefully adding capacity to groups of
output classes that the trained model itself considers similar. While
we demonstrate results on a computer vision task, this is not an assumption
underlying the approach, and we plan to extend it to other domains
in follow-up work.

In these experiments we have allocated a fixed extra capacity to each label
group, regardless of the number of labels in that group. Further investigation
is needed into strategies for the allocation of capacity to each label group.
Seemingly relevant factors include both the cardinality of each group and the
amount of training data available for the labels contained therein; however,
the difficulty of the discrimination task does not necessarily scale with
either of these.

In the case of the particular convolutional network we have described, it is not
obvious that the best place to connect these auxiliary stacks of hidden layers
is following the last convolutional layer. Most of the capacity, and therefore
arguably most of the discriminative knowledge in the network, is contained in
the fully connected layers, and appealing to this part of the network for
augmentation purposes seems natural. Nonetheless, it is possible that one or
more layers of group-specific convolutional feature maps could be beneficial as
well. Note that the augmentation procedure could also theoretically be applied
more than once, and not necessarily in the same location. Each subsequent
clustering and retraining step could potentially identify a complementary
division of the label space, capturing new information.

Finally, this can be seen as a small step towards the ``conditional
computation'' envisioned by \cite{bengio2013deep}, wherein relevant pathways of
a large network are conditionally activated based on task relevance. Here we
have focused on the relatively large gains to be had with computationally
inexpensive, targeted augmentations. Similar strategies could pave the way
towards networks with much higher capacity specialists that are only evaluated
when necessary.

\bibliography{iclr2015}

\begin{thebibliography}{10}
\providecommand{\natexlab}[1]{#1}
\providecommand{\url}[1]{\texttt{#1}}
\expandafter\ifx\csname urlstyle\endcsname\relax
  \providecommand{\doi}[1]{doi: #1}\else
  \providecommand{\doi}{doi: \begingroup \urlstyle{rm}\Url}\fi

\bibitem[Bengio(2013)]{bengio2013deep}
Bengio, Yoshua.
\newblock Deep learning of representations: Looking forward.
\newblock In \emph{Statistical Language and Speech Processing}, pp.\  1--37.
  Springer, 2013.

\bibitem[Bollacker et~al.(2008)Bollacker, Evans, Paritosh, Sturge, and
  Taylor]{Freebase}
Bollacker, Kurt, Evans, Colin, Paritosh, Praveen, Sturge, Tim, and Taylor,
  Jamie.
\newblock Freebase: a collaboratively created graph database for structuring
  human knowledge.
\newblock In \emph{Proceedings of the 2008 ACM SIGMOD international conference
  on Management of data}, pp.\  1247--1250. ACM, 2008.

\bibitem[Buciluǎ et~al.(2006)Buciluǎ, Caruana, and
  Niculescu-Mizil]{Bucilua2006}
Buciluǎ, Cristian, Caruana, Rich, and Niculescu-Mizil, Alexandru.
\newblock Model compression.
\newblock In \emph{Proceedings of the 12th ACM SIGKDD international conference
  on Knowledge discovery and data mining}, pp.\  535--541. ACM, 2006.

\bibitem[Chung(1997)]{chung1997spectral}
Chung, Fan~RK.
\newblock \emph{Spectral graph theory}, volume~92.
\newblock American Mathematical Soc., 1997.

\bibitem[Hinton et~al.(2014)Hinton, Vinyals, and Dean]{Hinton2014}
Hinton, Geoffrey~E., Vinyals, Oriol, and Dean, Jeff.
\newblock Distilling the knowledge in a neural network.
\newblock In \emph{NIPS 2014 Deep Learning Workshop}, 2014.

\bibitem[Jacobs et~al.(1991)Jacobs, Jordan, Nowlan, and
  Hinton]{jacobs1991adaptive}
Jacobs, Robert~A, Jordan, Michael~I, Nowlan, Steven~J, and Hinton, Geoffrey~E.
\newblock Adaptive mixtures of local experts.
\newblock \emph{Neural computation}, 3\penalty0 (1):\penalty0 79--87, 1991.

\bibitem[Krizhevsky et~al.(2012)Krizhevsky, Sutskever, and
  Hinton]{krizhevsky2012imagenet}
Krizhevsky, Alex, Sutskever, Ilya, and Hinton, Geoffrey~E.
\newblock Imagenet classification with deep convolutional neural networks.
\newblock In \emph{Advances in neural information processing systems}, pp.\
  1097--1105, 2012.

\bibitem[Ng et~al.(2002)Ng, Jordan, Weiss, et~al.]{Ng2002}
Ng, Andrew~Y, Jordan, Michael~I, Weiss, Yair, et~al.
\newblock On spectral clustering: Analysis and an algorithm.
\newblock \emph{Advances in neural information processing systems}, 2:\penalty0
  849--856, 2002.

\bibitem[Sermanet et~al.(2014)Sermanet, Eigen, Zhang, Mathieu, Fergus, and
  LeCun]{sermanet-iclr-14}
Sermanet, Pierre, Eigen, David, Zhang, Xiang, Mathieu, Michael, Fergus, Rob,
  and LeCun, Yann.
\newblock Overfeat: Integrated recognition, localization and detection using
  convolutional networks.
\newblock In \emph{International Conference on Learning Representations
  (ICLR2014)}. CBLS, April 2014.

\bibitem[{Szegedy} et~al.(2014){Szegedy}, {Liu}, {Jia}, {Sermanet}, {Reed},
  {Anguelov}, {Erhan}, {Vanhoucke}, and {Rabinovich}]{Szegedy14arXiv}
{Szegedy}, C., {Liu}, W., {Jia}, Y., {Sermanet}, P., {Reed}, S., {Anguelov},
  D., {Erhan}, D., {Vanhoucke}, V., and {Rabinovich}, A.
\newblock {Going Deeper with Convolutions}.
\newblock \emph{ArXiv e-prints}, September 2014.

\end{thebibliography}
\bibliographystyle{iclr2015}

\end{document}